\documentclass[runningheads,oribibl]{llncs}
\newcommand{\papertitle}{
    Verifying Low-dimensional Input Neural Networks via Input Quantization
}

\usepackage{mathtools}
\usepackage{amsmath}
\usepackage{amssymb}
\usepackage{multirow}
\usepackage{setspace}
\usepackage{threeparttable}
\usepackage{stfloats}
\usepackage{textcomp}
\usepackage[inline]{enumitem}
\usepackage{algorithm}
\usepackage{algpseudocode}
\usepackage{caption}
\usepackage{subcaption}
\usepackage{todonotes}
\usepackage{multirow}
\usepackage{bm}
\usepackage{bbm}
\usepackage{graphicx}
\usepackage[hidelinks, pdftex,
    pdfauthor={Kai Jia and Martin Rinard},
    pdftitle={\papertitle}]{hyperref}
\usepackage{url}
\usepackage{booktabs}       
\usepackage{amsfonts}       

\newcommand{\V}[1]{\bm{#1}}

\newcommand{\real}{\mathbb{R}}
\DeclareMathOperator{\argmax}{argmax}
\DeclareMathOperator{\argmin}{argmin}

\newcommand{\vx}{\V{x}}
\newcommand{\vyi}{\V{y_i}}
\newcommand{\vyhi}{\hat{\V{y_i}}}
\newcommand{\vxq}{\V{x}^q}
\newcommand{\vf}{\V{f}}
\newcommand{\vfq}{\V{f}^q}
\newcommand{\vq}{\V{q}}
\newcommand{\nth}[1]{#1^{\text{\hspace{.1em}th}}}
\newcommand{\pnorm}[2]{\left\| #2 \right\|_{#1}}

\newcommand{\defeq}{\vcentcolon=}

\usepackage[nameinlink,noabbrev]{cleveref}
\renewcommand{\cref}[1]{\Cref{#1}}
\newcommand{\eqnref}[1]{\hyperref[eqn:#1]{(\ref*{eqn:#1})}}

\usepackage[numbers,sort&compress]{natbib}

\makeatletter
\renewcommand\@biblabel[1]{#1.}
\makeatother

\begin{document}
\title{\papertitle}
%
%
\author{Kai Jia\inst{1}\and
Martin Rinard\inst{1}}
%
%
\institute{MIT CSAIL, Cambridge MA 02139, USA \\
    \email{\{jiakai,rinard\}@mit.edu}
}

\maketitle              

\begin{abstract}
    Deep neural networks are an attractive tool for compressing the control
    policy lookup tables in systems such as the Airborne Collision Avoidance
    System (ACAS). It is vital to ensure the safety of such neural controllers
    via verification techniques. The problem of analyzing ACAS Xu networks has
    motivated many successful neural network verifiers. These verifiers
    typically analyze the internal computation of neural networks to decide
    whether a property regarding the input/output holds. The intrinsic
    complexity of neural network computation renders such verifiers slow to run
    and vulnerable to floating-point error.

    This paper revisits the original problem of verifying ACAS Xu networks. The
    networks take low-dimensional sensory inputs with training data provided by
    a precomputed lookup table. We propose to prepend an input quantization
    layer to the network. Quantization allows efficient verification via input
    state enumeration, whose complexity is bounded by the size of the
    quantization space. Quantization is equivalent to nearest-neighbor
    interpolation at run time, which has been shown to provide acceptable
    accuracy for ACAS in simulation. Moreover, our technique can deliver exact
    verification results immune to floating-point error if we directly enumerate
    the network outputs on the target inference implementation or on an accurate
    simulation of the target implementation.

    \keywords{Neural network verification \and Verification by enumeration
        \and ACAS Xu network verification }
\end{abstract}


\section{Introduction}
\label{sec:intro}

The Airborne Collision Avoidance System (ACAS) is crucial for aircraft
safety~\citep{ kochenderfer2011robust}. This system aims to avoid collision with
intruding aircraft via automatically controlling the aircraft or advising a
human operator to take action. The ACAS typically takes low-dimensional sensory
inputs, including distance, direction, and speed for the intruder and ownship
aircraft, and provides a control policy which is a valuation for a set of
candidate actions such as ``weak left'' or ``strong right''. Recent work has
formulated aircraft dynamics under uncertainties such as advisory response delay
as a partially observable Markov decision process for which dynamic programming
can be used to compute values for different actions~\citep{
kochenderfer2015optimized}. The value function computed via dynamic programming
is often stored in a lookup table with millions of entries \citep{
kochenderfer2010decision} that require gigabytes of storage. While this table
could, in principle, be used to implement the ACAS, the high storage demand
makes it too costly to be embedded in practical flight control systems. This
situation has motivated the development of table compression techniques,
including block compression with reduced floating-point precision \citep{
kochenderfer2013compression} and decision trees \citep{julian2019deep}.

Recently, neural networks have emerged as an efficient alternative for
compressing the lookup tables in ACAS Xu (ACAS X for unmanned aircraft) by
approximating the value function with small neural networks. Specifically,
\citet{ julian2019deep} compresses the two-gigabyte lookup table into 45 neural
networks with 2.4MB of storage, where each network handles a partition of the
input space.

\citet{katz2017reluplex} proposes a set of safety properties for the ACAS Xu
networks, such as that a ``strong right'' advisory should be given when a nearby
intruder is approaching from the left. These safety properties have served as a
valuable benchmark to motivate and evaluate multiple verification algorithms
\citep{katz2017reluplex, wang2018formal, singh2019boosting, tran2020nnv,
bak2020improved}. Such verifiers typically need to perform exact or conservative
analysis of the internal neural network computation \citep{liu2019algorithms,
urban2021review}. Unfortunately, neural network verification is an NP-Complete
problem \citep{katz2017reluplex}, and therefore the verifiers need exponential
running time in the worst case and can be very slow in practice. In particular,
\citet{bak2020improved} recently presented the first verifier that is able to
analyze the properties $\phi_1$ to $\phi_4$ in the ACAS Xu benchmarks with a
time limit of 10 minutes for each case, but their verifier still needs 1.7 hours
to analyze the property $\phi_7$.

In summary, previous techniques perform the following steps to obtain and verify
their neural network controllers for ACAS:
\begin{enumerate}
    \item Compute a lookup table containing the scores of different actions
        given sensory states via dynamic programming.
    \item Train neural networks to approximate the lookup table.
    \item In deployed systems, use the neural networks to provide control
        advisories.
        \begin{itemize}
            \item At run time, the networks give interpolated scores for
                states not present in the original lookup table.
            \item Neural network verifiers that analyze the internal computing
                of neural networks are adopted to check if the networks meet
                certain safety specifications.
        \end{itemize}
\end{enumerate}

We propose instead to verify neural networks with low-dimensional inputs, such
as the ACAS Xu networks, via input quantization and state enumeration.
Specifically, we prepend a quantization layer to the network so that all the
internal computation is performed on the discretized input space. Our proposed
technique performs the following steps to obtain and verify a quantized neural
network:
\begin{enumerate}
    \item We take a pretrained network and prepend an input quantization layer
        to the network. The input quantization should be compatible with the
        original lookup table, i.e., preserving the grid points in the lookup
        table.
    \item In deployed systems, sensory inputs are first quantized by the input
        quantization layer. The original network then computes the scores for
        the quantized input.
        \begin{itemize}
            \item At run time, the quantization process is equivalent to
                nearest-neighbor interpolation.
            \item To verify the network for any specification, we enumerate all
                quantized states within the constraint of the specification and
                check if the network outputs meet the specification.
        \end{itemize}
\end{enumerate}

Our method provides the following desirable features:
\begin{enumerate}
    \item Our method  provides acceptable runtime accuracy for ACAS Xu. Our
        input quantization is equivalent to nearest-neighbor interpolation and
        gives identical results on the table grid points as the original
        continuous network. \citet{ julian2019deep} has shown that
        nearest-neighbor interpolation on the lookup table for runtime sensory
        inputs provides effective collision avoidance advisories in simulation.
    \item Our method enables efficient verification. Verifying the
        input-quantized networks for any safety specification takes nearly
        constant time bounded by evaluating the network on all the grid points
        in the quantized space. Multiple specifications can be verified
        simultaneously by evaluating the network on the grid once and checking
        the input and output conditions for each property. Our method provides a
        verification speedup of tens of thousands of times compared to the
        ReluVal~\citep{wang2018formal} verifier.
    \item Many existing verifiers do not accurately model floating-point
        arithmetic due to efficiency considerations, thus giving potentially
        incorrect verification results \citep{jia2020exploiting}. For example,
        \citet{ wang2018formal} reports that Reluplex \citep{katz2017reluplex}
        occasionally produces false adversarial examples due to floating-point
        error.

        By contrast, our verification result is exact (i.e., complete and sound)
        and does not suffer from floating-point error because we combine input
        quantization and complete enumeration of the effective input space.
        Moreover, input quantization allows directly verifying on the target
        implementation or an accurate simulation of the implementation, and
        therefore provides trustworthy safety guarantees for given neural
        network inference implementations.
    \item Our technique allows easily verifying more complicated network
        architectures, such as continuous-depth models \citep{chen2018neural}.
        Our verification only needs an efficient inference implementation for
        the networks. By contrast, extending other neural network verifiers to
        new network architectures requires significant effort.
\end{enumerate}

We recommend input quantization for neural networks with low-dimensional inputs
as long as the quantization provides sufficient accuracy for the target
application and the quantization space is small enough to allow efficient
enumeration. This technique enables efficient, exact, and robust verification
and provides reliable performance on the deployed platform.


\section{Method}

We formally describe our input-quantization method. This paper uses bold symbols
to represent vectors and regular symbols to represent scalars. The superscript
represents derived mathematical objects or exponentiation depending on the
context.

Let $\vf:\real^n \mapsto \real^m$ denote the computation of a neural network on
$n$-dimensional input space with $n$ being a small number. We propose to use
a quantized version of the network for both training and inference, defined as
\begin{align}
    \vfq(\vx) \defeq \vf(\vq(\vx))
\end{align}
where $\vq(\vx)$ is the quantization function such that $\vq(\vx)\in S$ with
$S$ being a finite-sized set. For a specification $\phi: \forall_{\vx} P(\vx)
\implies R(\vf(\vx))$ where $P(\cdot)$ and $R(\cdot)$ are predicates, we verify
$\phi$ regarding $\vfq$ by checking:
\begin{align}
    \phi^q \;:\quad & \forall \vxq \in S_p \implies R(\vf(\vxq)) \\
    \text{where } & S_p \defeq \{\vq(\vx) \;:\; P(\vx) \} \nonumber
\end{align}
Since $S_p \subseteq S$, the complexity of verifying $\phi^q$ is bounded by
$|S|$.

We quantize each dimension of $\vx$ independently via $\vq(\vx) = [q_1(x_1)
\ldots q_n(x_n)]$. Note that if some of the dimensions are highly correlated in
some application, we can quantize them together to avoid a complete Cartesian
product and thus reduce the size of the quantized space.

In many cases, the input space is uniformly quantized. Previous work has
utilized uniform input quantization for neural network verification \citep{
wu2020robustness, jia2020efficient} and uniform computation quantization for
efficient neural network inference \citep{ gholami2021survey}. Given a
quantization step $s_i$ and a bias value $b_i$, we define a uniform quantization
function $q_i(\cdot)$ as:
\begin{align}
    q_i(x_i) = \left\lfloor \frac{x_i - b_i}{s_i} \right\rceil s_i + b_i
    \label{eqn:uniform-quant}
\end{align}
where $\lfloor\cdot\rceil$ denotes rounding to the nearest integer.

The values of $q_i(\cdot)$ are essentially determined according to prior
knowledge about the target application and may thus be nonuniform. Let $Q_i =
\{v_i^1, \cdots, v_i^k\}$ denote the range of $q_i(\cdot)$. We use nearest
neighbor for nonuniform quantization:
\begin{align}
    q_i(x_i) = \argmin_{v_i^j} |v_i^j - x_i|
\end{align}

The ACAS Xu networks are trained on a lookup table $\V{L}: G \mapsto \real^m$,
where the domain $G\subset\real^n$ is a finite set. We choose the quantization
scheme so that the quantization preserves grid points, formally $\forall \vx \in
G: \vq(\vx) = \vx$. In this way, the training processes of $\vf(\cdot)$ and
$\vfq(\cdot)$ are identical. In fact, we directly prepend $\vq(\cdot)$ as an
input quantization layer to a pretrained network $\vf(\cdot)$ to obtain
$\vfq(\cdot)$. Note that we can use a denser quantization than the grid points
in $G$ so that prediction accuracy might get improved by using the neural
network as an interpolator.


\section{Experiments}

We evaluate our method on checking the safety properties for the ACAS Xu
networks \citep{katz2017reluplex}. Note that accuracy of input-quantized
networks in deployed systems is acceptable since the quantization is equivalent
to nearest-neighbor interpolation that has been shown to provide effective
collision avoidance advisories in simulation \citep{julian2019deep}.

Experiments in this section focus on evaluating the runtime overhead of input
quantization and the actual speed of verification by enumerating quantized
states. We train two networks of different sizes to evaluate the scalability of
the proposed method.

\subsection{Experimental Setup}

The horizontal CAS problem takes seven inputs as described in
\cref{tab:input-def}, and generates one of the five possible
advisories: COC (clear of conflict), WL (weak left), WR (weak right), SL (strong
left), and SR (strong right).

\begin{table}[t]
    \begin{threeparttable}
    \centering
    \scriptsize
    \caption{
        Description of horizontal CAS inputs. The last column describes the
        values used to generate the lookup table, which are taken from the
        open-source implementation of HorizontalCAS \citep{
        julian2019guaranteeing} and the Appendix VI of \citet{
        katz2017reluplex}.
        \label{tab:input-def}
    }
    \vskip .5em
    \begin{tabular}{lll}
        \toprule
        Symbol & Description & Values in the lookup table \\
        \midrule
        $\rho$ (m) & Distance from ownship to intruder &
            32 values between 0 and 56000 \tnote{1}\\
        $\theta$ (rad) & Angle to intruder \tnote{2}
            & 41 evenly spaced values between $-\pi$ and $\pi$ \\
        $\psi$ (rad) & Heading angle of intruder \tnote{2}
            & 41 evenly spaced values between $-\pi$ and $\pi$ \\
        $v_{own}$ (m/s) & Speed of ownship & $\{50, 100, 150, 200\}$ \\
        $v_{int}$ (m/s) & Speed of intruder & $\{50, 100, 150, 200\}$ \\
        $\tau$ (sec) & Time until loss of vertical separation &
            $\{0, 1, 5, 10, 20, 40, 60\}$ \\
        $\alpha_{prev}$ & Previous advisory & \{COC, WL, WR, SL, SR\} \\
        \bottomrule
    \end{tabular}
    \vspace{1ex}
    \begin{tablenotes}
        \item[1] Distance values are nonuniformly distributed. They are given in
            the source code of \citet{julian2019guaranteeing}:
            \url{https://github.com/sisl/HorizontalCAS/blob/cd72ffc073240bcd4f0eb9164f441d3ad3fdc074/GenerateTable/mdp/constants.jl\#L19}
        \item[2] Angle is measured relative to ownship heading direction.
    \end{tablenotes}
    \end{threeparttable}
\end{table}

\citet{julian2016policy} proposes to train a collection of neural networks where
each network works with a pair of specific $(\tau,\, \alpha_{prev})$ values,
takes the remaining five values as network inputs, and approximates the
corresponding scores in the lookup table. Although ACAS only needs to suggest
the action with the maximal score, the network is still trained to approximate
the original scores in the table instead of directly giving the best action
because the numerical scores are used in a Kalman filter to improve system
robustness in the face of state measurement uncertainty~\cite{
julian2016policy}. In order to maintain the action recommendation of the
original table while reducing score approximation error, \citet{julian2016policy}
adopts an asymmetric loss function that imposes a higher penalty if the network
and the lookup table give different action advisories.

\citet{katz2017reluplex} proposes a few ACAS Xu safety properties as a sanity
check for the networks trained by \citet{julian2016policy}. These properties
have also served as a useful benchmark for many neural network verifiers.
Although the pretrained networks of \citet{julian2016policy} are publicly
accessible, the authors told us that they could not provide the training data or
the source code due to regulatory reasons. They suggested that we use their
open-source HorizontalCAS system \citep{ julian2019guaranteeing} to generate the
lookup tables to train our own networks. However, HorizontalCAS networks differ
from the original ACAS Xu networks in that they only have three inputs by fixing
$v_{own}=200$ and $v_{int}=185$. We modified the source code of HorizontalCAS to
match the input description in \cref{tab:input-def} so that we can directly use
the ReluVal \citep{wang2018formal} verifier.

We evaluate our method by analyzing the property $\phi_9$ proposed in
\citet{katz2017reluplex}, which usually takes the longest time to verify among
all the properties for many verifiers~\cite{katz2017reluplex, wang2018formal,
singh2018robustness}. Other properties share a similar form but have different
input constraints and output requirements. Note that property $\phi_9$ is the
most compatible with the open-source HorizontalCAS because the input constraints
of other properties are beyond the ranges in \cref{tab:input-def}. For example,
property $\phi_1$ has $v_{own} \geq 1145$ but the quantization scheme of
$v_{own}$ for the original ACAS Xu networks is not publicly available.

The specification of $\phi_9$ is:
\begin{itemize}
    \item {\bf Description:} Even if the previous advisory was ``weak right'',
        the presence of a nearby intruder will cause the network to output a
        ``strong left'' advisory instead.
    \item {\bf Tested on:} the network trained on $\tau=5$ and
        $\alpha_{prev}=\text{WR}$
    \item {\bf Input constraints:} $2000 \le \rho \le 7000$,
        $-0.4 \le \theta \le -0.14$, $-3.141592 \le \psi \le -3.141592 + 0.01$,
        $100 \le v_{own} \le 150$, $0 \le v_{int} \le 150$.
\end{itemize}

We conduct the experiments on a workstation equipped with two GPUs (NVIDIA Titan
RTX and NVIDIA GeForce RTX 2070 SUPER), 128 GiB of RAM, and an AMD Ryzen
Threadripper 2970WX processor. We train two neural networks for property
$\phi_9$ (i.e., with $\tau=5$ and $\alpha_{prev}=\text{WR}$) with PyTorch.

Our small network has five hidden layers with 50 neurons in each layer, and our
large network has seven hidden layers with 100 neurons in each layer. We use the
ReLU activation.

We implement the nearest-neighbor quantization for $\rho$ via directly indexing
a lookup table. The greatest common divisor of differences between adjacent
quantized $\rho$ values is 5. Therefore, we precompute a lookup table $\V{U}$
such that $U_{\lfloor \rho/5 \rceil}$ is the nearest neighbor of $\rho$ in the
set of quantized values. We use the \verb|torch.index_select| operator provided
by PyTorch to take elements in the lookup table in a batched manner. Other
network inputs use uniform quantization as described in \cref{tab:input-def}. We
implement uniform quantization according to the equation \eqnref{uniform-quant}.

\subsection{Experimental Results}

\begin{table}[t]
    \centering
    \caption{Accuracies achieved by the networks evaluated on the lookup table.
        For comparison, \citet{julian2019guaranteeing} reports an accuracy of
        97.9\% for networks trained only with three out of the five inputs (they
        fixed $v_{own}=200$ and $v_{int}=185$). This table shows that our
        network achieves sufficient accuracy for practical use.
        \label{tab:acc}
    }
    \vskip .5em
    \begin{tabular}{lrr}
        \toprule
        Metric          & Small network  & \hspace{1em} Large network   \\
        \midrule
        Policy accuracy         & 96.87\%         & 98.54\%         \\
        Score $\ell_1$ error    & 0.052           & 0.026           \\
        Score $\ell_2$ error    & $1.3\times10^{-3}$ & $3.3\times10^{-4}$  \\
        \bottomrule
    \end{tabular}
\end{table}

\begin{table}[t]
    \caption{Comparing verification time (in seconds) for the property $\phi_9$
        on two methods: the ReluVal verifier~\cite{wang2018formal} that runs on
        multiple cores, and exhaustive enumeration in the quantized input space
        on a single CPU core. This table shows that verification by enumerating
        quantized input states is significantly faster in our case and also more
        scalable regarding different network sizes.
        \label{tab:cas/verify-time}
    }
    \renewcommand{\TPTminimum}{\linewidth}
    \begin{threeparttable}
    \vskip .5em
    \makebox[\linewidth]{
    \begin{tabular}{lrr}
        \toprule
        Verification method  & Small network & \hspace{1em} Large network \\
        \midrule
        ReluVal \citep{wang2018formal} & 0.622 & 171.239 \\
        Input quantization - specific \tnote{1} &
            0.002 & 0.002 \\
        Input quantization - all \tnote{2} & 0.384 & 0.866  \\
        \bottomrule
    \end{tabular}
    }
    \begin{tablenotes}
        \item[1] Network is evaluated on the 60 input states that fall within
            the input constraint of $\phi_9$.
        \item[2] Network is evaluated on all the 860,672 input states in a
            batched manner. This time is the upper bound for verifying any
            first-order specification in the form of $\forall_{\vx}P(\vx)
            \implies R(\vf(\vx))$ by ignoring the time on evaluating predicates
            $P(\cdot)$ and $R(\cdot)$.
    \end{tablenotes}
    \end{threeparttable}
\end{table}

Let $\vyi\in\real^5$ (resp. $\vyhi\in\real^5$) denote the scores given by the
network (resp. the original lookup table) for the five candidate actions on the
$\nth{i} $ lookup table entry. We consider three accuracy measurements, assuming
a uniform distribution of the table index $i$:
\begin{itemize}
    \item \emph{Policy accuracy} is the probability that the network recommends
        the same action as the original lookup table. Formally, it is $P(\argmax
        \vyi = \argmax \vyhi)$.
    \item \emph{Score $\ell_1$ error} measures the $\ell_1$ error of
        approximated scores, defined as $E(\pnorm{1}{\vyi - \vyhi})$,
        where $\pnorm{1}{\vx} \defeq \sum_i |x_i|$.
    \item \emph{Score $\ell_2$ error} measures the $\ell_2$ error of
        approximated scores, defined as $E(\pnorm{2}{\vyi - \vyhi})$,
        where $\pnorm{2}{\vx} \defeq \sqrt{\sum_i x_i^2}$.
\end{itemize}

\cref{tab:acc} presents the accuracies achieved by our networks, which shows
that our training achieves comparable results as the HorizontalCAS system
\citep{julian2019guaranteeing}.

To verify the networks, we prepend them with an input quantization layer that
implements the quantization scheme given in \cref{tab:input-def}. To verify any
specification or a set of specifications, we evaluate the network on all the 860,
672 points in the quantized space and check if each input/output pair meets the
specification(s). Evaluating the network on the grid points takes 0.384 seconds
for the small network and 0.866 seconds for the large one. We evaluate the
network on multiple inputs in a batched manner to benefit from optimized
numerical computing routines included in PyTorch. Adding the quantization layer
incurs about 2\% runtime overhead. We do not do any performance engineering and
use the off-the-shelf implementation provided by PyTorch. Our verification speed
can be further improved by using multiple CPU cores or using the GPU.

We also compare our method with ReluVal \citep{wang2018formal} on verifying the
property $\phi_9$. The input constraint of $\phi_9$ consists of only 60 states
in the quantized space. Therefore, we only need to check if the network
constantly gives the ``weak right' advisory for all the 60 states to verify
$\phi_9$. As shown in \cref{tab:cas/verify-time}, input quantization
significantly reduces the verification time compared to the ReluVal solver.


\section{Conclusion}

This paper advocates input quantization for the verification of neural networks
with low-dimensional inputs. Our experiments show that this technique is
significantly faster and more scalable than verifiers that analyze the internal
computations of the neural networks on verifying ACAS Xu networks. Moreover, our
method does not suffer from the floating-point discrepancy between the verifier
and the network inference implementation. In general, our method applies to
deterministic floating-point programs that take low-dimensional inputs as long
as the target application tolerates input quantization such that enumerating all
the quantized values takes acceptable time.


%
%
\bibliographystyle{splncs04nat}
\bibliography{references}
\end{document}